\documentclass[final]{cvpr}

\usepackage{times}
\usepackage{epsfig}
\usepackage{graphicx}
\usepackage{amsmath}
\usepackage{amssymb}
\usepackage{gensymb}
\usepackage{bbm}
\usepackage{array}
\usepackage{makecell}
\usepackage{amsfonts}
\usepackage{booktabs}
\usepackage{siunitx}
\usepackage{multirow}
\usepackage[font=small]{caption}
\usepackage{mathtools}
\usepackage{algpseudocode}
\usepackage{algorithm}
\usepackage{algorithmicx}
\usepackage{etoolbox}
\usepackage{xcolor}

\newcommand{\cL}{\mathcal{L}}

\DeclareMathOperator*{\argmax}{arg\,max}
\DeclareMathOperator*{\argmin}{arg\,min}
\usepackage{enumitem}

\usepackage[pagebackref=true,breaklinks=true,colorlinks,bookmarks=false]{hyperref}

\def\cvprPaperID{6732} 
\def\confYear{CVPR 2021}

\begin{document}

\title{Boosting the Performance of Semi-Supervised Learning \\ with Unsupervised Clustering}

\author{Boaz Lerner\\
The Hebrew University of Jerusalem\\
{\tt\small boaz.lerner@mail.huji.ac.il}
\and
Guy Shiran\\
The Hebrew University of Jerusalem\\
{\tt\small guy.shiran@mail.huji.ac.il}

\and
Daphna Weinshall\\
The Hebrew University of Jerusalem\\
{\tt\small daphna@cs.huji.ac.il}
}

\maketitle

\begin{abstract}
   
Recently, Semi-Supervised Learning (SSL) has shown much promise in leveraging unlabeled data while being provided with very few labels. In this paper, we show that ignoring the labels altogether for whole epochs intermittently during training can significantly improve performance in the small sample regime. More specifically, we propose to train a network on two tasks jointly. The primary classification task is exposed to both the unlabeled and the scarcely annotated data, whereas the secondary task seeks to cluster the data without any labels. As opposed to hand-crafted pretext tasks frequently used in self-supervision, our clustering phase utilizes the same classification network and head in an attempt to relax the primary task and propagate the information from the labels without overfitting them. On top of that, the self-supervised technique of classifying image rotations is incorporated during the unsupervised learning phase to stabilize training. We demonstrate our method's efficacy in boosting several state-of-the-art SSL algorithms, significantly improving their results and reducing running time in various standard semi-supervised benchmarks, including 92.6\% accuracy on CIFAR-10 and 96.9\% on SVHN, using only 4 labels per class in each task. We also notably improve the results in the extreme cases of 1,2 and 3 labels per class, and show that features learned by our model are more meaningful for separating the data. 

\end{abstract}

\section{Introduction}
In recent years we have seen huge improvement in the performance of deep learning methods in various computer vision tasks. However, most models require large amounts of annotated data. Collecting this data is a tedious and expensive process. Moreover, learning from so many labels is very different from the way that we as humans learn, and from the way we perceive intelligence. Therefore, it seems desirable, in our journey towards strong AI, to develop models that rely less on data annotated by humans, and are capable of extracting features in an unsupervised manner.  

\begin{figure}[t]
\begin{center}
\includegraphics[width=1.0\columnwidth]{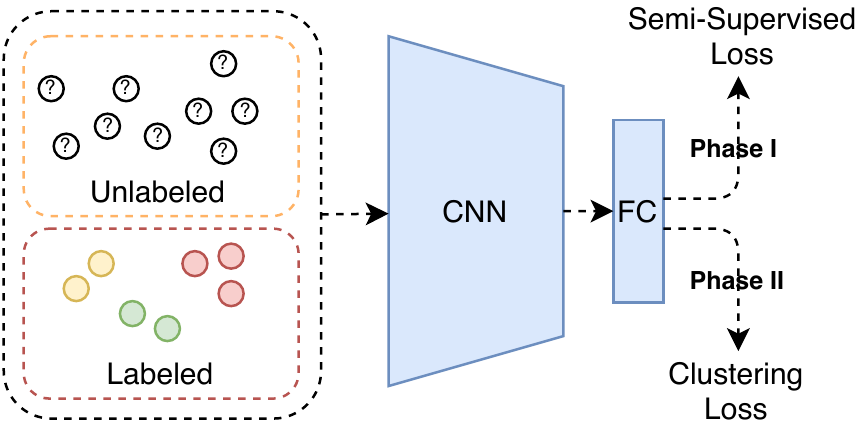}
 \hfill
    \caption{Our method iterates between two phases. In the first phase we train our model with the available labeled and unlabeled data points using a semi-supervised learning algorithm. In the second phase we train the same model to cluster all the data points, without using any labels.}
    \label{fig: method_diagram}
    \vspace{-0.5cm}
\end{center}
\end{figure}

Semi-Supervised Learning (SSL) is an attempt to tackle this issue and bridge the gap between supervised and unsupervised learning. The typical setting of the problem is that we are given a small amount of labeled data and a possibly large amount of unlabeled data, where both are sampled from the same or similar distributions. Most techniques derive an objective which is split into two separate terms for labeled and unlabeled data \cite{s4l, uda}. In this way, every gradient step is influenced by the labels. \cite{scarce_annotations} is unusual in that respect, as their method iterates between supervised learning with only labeled data, and unsupervised learning with only unlabeled data. However, their unsupervised stage relies exclusively on fixed pseudo-labels obtained in the supervised stage. Hence, this stage can be considered 'supervised learning with noisy labels'.

In this paper, we wish to benefit from real unsupervised learning, undistracted by a small number of possibly uncharacteristic training points, within the SSL framework. The approach is illustrated in Fig.~\ref{fig: method_diagram}. We start by describing a clustering algorithm that can be easily integrated with any deep SSL method.  This algorithm serves as a secondary task to the principal classification task. Unlike \cite{scarce_annotations}, the pseudo labels (targets) may be propagated from the real labels seen during classification, but they are likely to change during this clustering phase. Also differing from some self-supervised auxiliary tasks used in \cite{remixmatch, s4l}, we use the exact same architecture (network and head) for clustering. Our main goal is to add to the learning protocol a phase which does not depend on the labels. When learning the secondary task, the goal is to separate the data into clusters without assigning names to those clusters. At the same time, self-supervision \cite{rotnet} is used to stabilize and accelerate training during the clustering phase.

Our main contributions in this paper are the following:
\begin{itemize}[itemsep=1pt, leftmargin=5mm]
  \item Devise a new approach for semi-supervised learning, which is based on solving an unsupervised clustering task together with a classification task.  
  \item Show how realizing this approach by integrating a new deep clustering algorithm with three existing SSL algorithms can boost their performance and surpass state-of-the-art results on 3 benchmark datasets.   
\end{itemize}


\section{Related Work}
In this work we employ several methods from various fields, including deep clustering, self-supervision and semi-supervised learning. We therefore review the most significant recent developments in each of those fields.    


\subsection{Deep Clustering}
The task of unsupervised clustering is a long-standing problem and a highly challenging one, especially when it meets the high-dimensionality of images. Classical algorithms, such as k-means \cite{k-means} and Gaussian Mixture Models (GMM) \cite{GMM}, struggle in this domain as the raw data is not very informative, and thus the need for succinct and meaningful representation of images is critical. In recent years, deep clustering frameworks have become increasingly competent in solving this task. Typically, these models  jointly learn image features alongside cluster assignments by training a deep network with some clustering loss in an end-to-end fashion. The coupling of learning image features and clusters together allows the deep network to better adapt its image features to the task of clustering. 

In Deep Embedded Clustering (DEC) \cite{dec}, an encoder network is first initialized by pre-training on a reconstruction task alongside a decoder (i.e an auto-encoder) which is afterwards discarded. Then, cluster centroids in the embedding space are iteratively computed and refined until convergence. Joint Unsupervised Learning (JULE) \cite{jule} trains a model in an end-to-end fashion by iteratively merging clusters of deep representations and updating the network's parameters in a hierarchical fashion. Recently, Invariant Information Clustering (IIC) \cite{iic} proposed a novel information-theory approach for clustering, in which they maximize the mutual information between deep embeddings obtained by two different random augmentations applied to the same image, and rely on the natural characteristics of the mutual information loss to produce a clustering of the data. 


\subsection{Self-Supervised Learning}

Self-supervised learning is an approach to learning in an unsupervised manner by solving a pretext supervised task. The supervisory signals are gathered automatically from the data without the need for manual labeling. The task is designed in a way that implicitly requires learning of useful image representation, e.g. predicting the relative position of patches in an image \cite{relative_position_of_patches_prediction}, or solving jigsaw puzzles \cite{jigsaw_puzzle}. A more recent method \cite{rotnet} predicts image rotations (RotNet) and has also been used as an auxiliary task to stabilize and improve training in semi-supervised \cite{s4l} and image generation tasks \cite{self_supervised_gans}. Similalry, our method also employs RotNet for this purpose. 


\subsection{Semi-Supervised Learning}

Semi-supervised learning (SSL) refers to a family of algorithms aimed at learning from both labeled and unlabeled data. Thanks to the relative simplicity of collecting big unlabeled datasets without manual labeling, the field has seen an increasing interest in the last few years. As a result, the performance gap between supervised and unsupervised methods has been consistently diminishing, with the number of labels used to achieve comparable results getting considerably smaller. Given the vast amount of techniques proposed in the literature, we review only the latest and most influential works using deep networks, and refer the reader to \cite{SSL_book} for a comprehensive survey.  

Most deep learning approaches revolve around the two fundamental concepts of entropy minimization and consistency regularization. An intuitive assumption is that the decision boundaries of a classifier should not pass in highly dense areas. Minimizing the entropy of a model's prediction on unlabeled data is a common approach to facilitate this heuristic. It can be done explicitly \cite{entropy_minimization} or quite often implicitly by psuedo-labeling \cite{pseudo-label, scarce_annotations, fixmatch}, a method that assigns an artificial label to an unlabeled image and trains the network to predict that label. Consistency regularization refers to the assumption that small perturbations to the data should not affect its semantics, and hence the label. It is reasonable then to force the model to output consistent predictions to all perturbed versions of the same sample. Consequently, a lot of recent research makes use of complex augmentation strategies \cite{auto_augment, rand_augment, remixmatch, uda}. FixMatch \cite{fixmatch}, which is the most relevant to our work, combines both approaches and will be thoroughly discussed in Section~\ref{subsec: fixmatch}.

\section{Proposed Method}
\label{sec:method}

In this section we present an SSL algorithm that alternates between two phases: unsupervised clustering and semi-supervised classification. The method is designed especially to perform well in the very-few-labels scenario. 
 
As the SSL building block we use in our experiments a variety of recent state-of-the-art algorithms, including FixMatch~\cite{fixmatch} which is used in most of the experiments, as well as MixMatch~\cite{mixmatch} and UDA~\cite{uda}. Our main goal is to equip them with an additional unsupervised mechanism, in order to reduce their susceptibility to outliers in the small labeled sample. To this end we describe an effective unsupervised deep clustering method. This building block can also be replaced in order to improve performance. 

We start by describing each component separately, and then describe their integration into a single coherent model.

\subsection{Unsupervised Clustering}

In a typical supervised classification setting we are given pairs of images and labels $\{x_i,y_i\}_{i=1}^n$, and train a parameterized model $f_\theta$ by solving:
\begin{equation}
\min_{\theta} \frac{1}{n}\sum_{i=1}^n \ell (f_\theta (x_i), y_i),
\end{equation}
where $\ell$ is some loss function. In the setting of unsupervised clustering, where ground truth labels $y_1,...,y_n$ are not available, we can attempt to learn them alongside the model's parameters:
\begin{equation}
\min_{\theta, y_1,...,y_n} \frac{1}{n}\sum_{i=1}^n \ell (f_\theta (x_i), y_i).
\end{equation}

Without additional constraints, this optimization procedure is prone to suffer from mode collapse, where all images are assigned the same label $y_1=\cdots=y_n$. To overcome this susceptibility, we add a constraint to the optimization formulation that explicitly prevents this from happening:
\begin{equation}
\label{equation: constraint}
\begin{gathered}
\forall k\in [K] \quad \sum_{i=1}^n 1_{y_i=k}\geq\alpha\frac{n}{K}, \quad 0<\alpha\leq1.
\end{gathered}
\end{equation}
Here $K$ denotes the number of classes in the dataset and $\alpha$ is a hyper-parameter. Eq.~\ref{equation: constraint} guarantees that each cluster has a minimal number of images assigned to it.

In order to solve this optimization problem, we adapt a similar approach to the one taken in \cite{nat}, where the task of representation learning is addressed. In \cite{nat}, the problem of feature collapse is tackled by fixing static feature vectors $\{y_i\}_{i=1}^n$ at the beginning of training. Throughout the training, the algorithm learns the model's parameters and a one-to-one mapping $P:[n]\rightarrow[n]$ from images $\{x_i\}_{i=1}^n$ to those fixed targets.

In our case, we are interested in clustering the data, and hence it is reasonable to set the targets to be one-hot vectors in $\mathbb{R}^k$, which we denote by $e_1,\ldots,e_k$. To enforce (\ref{equation: constraint}), the targets are set to $T=\{y_i|\forall k\in [K], \sum_{i=1}^n \mathbbm{1}_{y_i=e_k}=\alpha\frac{n}{K}\}$, with $\alpha\frac{n}{K}$ target instances per cluster. Before training begins, each target is randomly assigned to an image in the dataset. Note that some images may not be paired up with a target as there might be more images than targets. 

The optimization problem is solved stochastically one mini-batch at a time, where each iteration consists of two steps. Given a mini-batch $X_b=\{x_1',\ldots,x_b'\} \subseteq X$  of $b$ images and their current targets $T_c=\{t_1',\ldots,t_c'\} \subseteq T$ ($b\geq c$), the first step finds the best assignment of targets to images, denoted by $P^*:[c]\rightarrow[b]$, while the network's parameters are kept fixed. This is accomplished by minimizing the following objective with the Hungarian method~\cite{hungarian}: 
\begin{equation}\label{equation: generalized permutation objective}
P^* = \argmin_{P} \sum_{i=1}^{c} ||f_\theta(x'_{P(i)}) - t'_i||_2^2.
\end{equation}

Recall that not all images are necessarily assigned a target. Among the unassigned images, only those with confidence exceeding a certain threshold are assigned to pseudo-targets. A similar approach is adopted in \cite{fixmatch, Chang2017DeepAI}. For an unassigned image $x'_k$ to be considered confident, we require that $||f_\theta(x'_{k}) - e_{\argmax f_\theta(x'_k)}||_2^2 < \rho$, where $\rho$ is a hyper-parameter. In this case, the pseudo-target assigned to $x'_k$ is $\tilde{y}_k = e_{\argmax f_\theta(x'_k)}$. 

In the second step of the optimization scheme, we update the model's parameters with a gradient step, which minimizes the distance of the model's outputs from the targets or pseudo-targets found in the first step. Specifically, if we denote the unassigned confident images by $C$, and the batch images being processed by $S = Im(P^*) \cup C$, then in the second step each image in $S$ is augmented $r$ times with a stochastic function $g$, where all $r$ versions of the same image are matched to the same target. This is formulated as minimizing the following objective w.r.t. $\theta$:
\begin{equation}\label{equation: generalized total objective}
\cL_c = \frac{1}{|S|r}\sum_{i \in S}\sum_{j=1}^r
||f_\theta(g(x'_i)) - \tilde{y}_i)||_2^2, 
\end{equation}
where $\tilde{y}_i$ is either the target or the pseudo-target of image $x'_i$. The objective is minimized via stochastic gradient descent. Note that unassigned images with low confidence are ignored and do not influence the optimization. As in \cite{nat}, we implement $f_\theta$ as a ConvNet and normalize its output such that $||f_\theta(x)||_2=1$.

To further enhance the representation capabilities of the model, which may in turn facilitate better clustering, we train the same model on an additional auxiliary task. Specifically, we employ the self-supervised task of predicting image rotations (RotNet) \cite{rotnet}, as it has a proven record of efficiently improving ConvNets representations in a variety of tasks \cite{s4l, self_supervised_gans, few_shot_self_supervision}. We do this by feeding the penultimate layer of $f_\theta$ into another head, which is used to generate the rotation predictions for the RotNet task.

The full clustering algorithm is detailed below in Alg.~\ref{algo: clustering algorithm}.

\begin{algorithm}[ht] 
            \caption{Unsupervised Clustering} 
            \begin{flushleft}
            
                \textbf{INPUT:} $X = \{x_i\}_{i=1}^n$ - unlabeled dataset \\
                \hspace{33pt} $f_\theta$ - convnet with two heads and parameters $\theta$ \\
                \hspace{33pt} $K$ - number of clusters \\
                \hspace{33pt} $g$ - stochastic augmentation function \\
                \hspace{33pt} $\lambda_i$ - learning rate at epoch $i$ \\
                \hspace{33pt} $r$ - number of times $g$ is applied to an image \\
                \hspace{33pt} $\alpha$ - ratio of dataset that will have  targets \\
                \hspace{33pt} $\rho$ - maximal distance to be considered confident \\
            \end{flushleft}
            \begin{algorithmic}
            
            \State $T \leftarrow [\,]$ 
            \For {$k$=1 to $K$}
            \Comment{initialize targets}
            \For {$i$=1 to $\alpha \frac{n}{K}$} 
            \State append $e_k$ to $T$
            \Comment{$e_k$ is the $k$th unit vector in $\mathbb{R}^K$}
            \EndFor
            \EndFor
            \State $\forall i \in [\alpha n] \quad A(x_i) = T[i]$ 
            \Comment {initialize assignments}

            \For {\textit{i=1...epochs}}
            \For {\textit{j=1...iters}}
            
            \State sample a batch $(\{x'_i\}_{i=1}^b, \{t'_i\}_{i=1}^c)$
            \Comment{$b\geq c$}
            \State compute $P^*$ with Eq.~\ref{equation: generalized permutation objective}
            \State  $\forall i \in [c] \quad A(x_{P^*(i)}) = t'_{i}$
            \Comment{update assignments}
            \State update the parameters with gradient step of Eq.~\ref{equation: generalized total objective}: 
            \State $$ \theta \leftarrow \theta - \lambda_i \nabla_{\theta} \cL_c $$ 

            \EndFor
            \For {\textit{j=1...iters}}
            \State sample a batch $X_b$
            \State $\forall d \in  \{0^{\circ}, 90^{\circ}, 180^{\circ}, 270^{\circ}\}$, rotate $X_b$ $d$ degrees
            \State update parameters with a gradient step of the \State cross-entropy loss $\cL_r$:
            
            $$ \theta \leftarrow \theta - \lambda_i \nabla_{\theta} \cL_r $$
            
            \EndFor
            \EndFor

        \end{algorithmic}
        \label{algo: clustering algorithm}
\end{algorithm}


\subsection{Semi-Supervised Classification} \label{subsec: fixmatch}

Semi-supervised classification is carried out in most of our experiments by adopting the FixMatch method, as it currently yields state-of-the-art results when relying on a small labeled sample. FixMatch combines two heuristics that are commonly used in SSL, \emph{consistency regularization} and \emph{pseudo-labelling}. These two heuristics are expressed as part of the loss function applied to unlabeled data during the training of a neural network, while labeled data is used to optimize the standard cross-entropy loss. 

Formally, given a batch of $b$ images $X = \{x_1,\ldots,x_b\}$ and their labels $Y = \{y_1,\ldots,y_b\}$, and another batch of $b'$ unlabeled images $U = \{u_1,\ldots,u_{b'}\}$, FixMatch predicts the class distribution of the network's output on a weakly-augmented expansion of the unlabeled batch, and uses these predictions as hard pseudo-labels for a strongly-augmented expansion of the same images. Thus, if we denote the network by $f_\theta$ and the stochastic weak and strong augmentation functions by $g$ and $q$ respectively, the pseudo-label of image $u_i$ becomes $y'_i = \max(f_\theta(g(x_i))$, and the loss term on the unlabeled batch can be written as:
\begin{equation*}
\cL_u = \frac{1}{|\{ i \in [b'] \mid y'_i \geq \tau\}|}\sum_{i=1}^{b'}\mathbbm{1}(y'_i \geq \tau)H(f_\theta(q(u_i)),y'_i).
\end{equation*}
Above, $H$ denotes the cross-entropy loss and $\tau$ denotes a hyper-parameter that determines the threshold confidence above which the image will be considered in the update of the network's parameters (similar to $\rho$ defined above). The loss on labeled data is simply:
\begin{equation}\label{equation: fixmatch_labeled_loss}
\cL_s = \frac{1}{b}\sum_{i=1}^{b}H(f_\theta(g(x_i)),y_i),
\end{equation}
and the total loss is a combination of them both: $\cL_s + \lambda_u\cL_u$, where $\lambda_u$ denotes a hyper-parameter balancing the weights of the two terms.  

The weak augmentations used in the algorithm include the standard flip and shift transformations. First, the images are flipped horizontally with 50\% probability, and then they are randomly translated by up to 12.5\% vertically and horizontally. As strong augmentations, two variants of AutoAugment~\cite{auto_augment} are used, RandAugment~\cite{rand_augment} and CTAugment~\cite{remixmatch}. Both are followed by Cutout~\cite{cutout}.

\subsection{Integrated Method}

The basic idea underlying our method is that if the number of labels is small, it benefits a classification algorithm to occasionally refrain from taking the labels into account. To achieve this goal, our method alternates traditional semi-supervised training epochs with full epochs that optimize the unsupervised loss in (\ref{equation: generalized total objective}) while ignoring the labels. This design aims to learn meaningful features, which can compensate for the shortage of labels and help the model generalize better. As a result, the model is less susceptible to overfitting the few labeled datapoints, especially when they do not agree well with the total data distribution. 

Specifically, we use the same network architecture and weights to solve the two different tasks described above jointly. To this end the algorithm alternates between FixMatch epochs and clustering epochs. We also perform several RotNet warmup epochs, as this has proven useful for accelerating the learning.  

Given the information propagated from the labels during the semi-supervised phase, clustering can be seen as a surprisingly easier task that attempts to separate the data without giving names to the different clusters thus created. Along the way, the mini-batch permutation optimization gives the network a chance to swiftly switch the targets of images whose confidence level is too low. Then, in the next supervised phase, the algorithm tries again to give those clusters names and refine the boundaries between them. This cycle is repeated until convergence.

Alg.~\ref{algo: complete algorithm} summarizes the method. We make the code available in the Supplementary Material. In the next section, we show its effectiveness in semi-supervised learning with a small labeled sample. In these experiments, the semi-supervised step is realized with more out-of-the-box SSL algorithms for comparison.   
\begin{algorithm}[t] 
            \caption{Boosted SSL} 
            \begin{flushleft}
            
                \textbf{INPUT:} $U=\{u_i\}_{i=1}^n$ - unlabeled dataset \\
                \hspace{33pt} $(X, Y)=\{x_i,y_i\}_{i=1}^m$ - labeled dataset \\
                \hspace{33pt} $SSL\_ALGO$ - some deep SSL algorithm \\
                \hspace{33pt} $C\_ALGO$ - our clustering algorithm from \ref{algo: clustering algorithm} \\
                \hspace{33pt} $f_\theta$ - convnet with two heads and parameters $\theta$ \\
                \hspace{33pt} $g$ - stochastic augmentation function \\
                \hspace{33pt} $\lambda_{i,j}$ - learning rate at iteration $i$, epoch $j$ \\
                \hspace{33pt} $r$ - number of times $g$ is applied to an image \\
                \hspace{33pt} $\alpha$ - ratio of dataset that will have targets \\
                \hspace{33pt} $\rho$ - maximal distance to be considered confident \\
            \end{flushleft}
            \begin{algorithmic}

            \For {\textit{i=1...iters}}
            \For {\textit{j=1...$e_1$}}
            \State run $SSL\_ALGO(X, Y, U, f_\theta)$ for one epoch
            
            \EndFor
            
            \For {\textit{j=1...$e_2$}}
            \State run $C\_\,ALGO(U, c, f_\theta, g, \lambda_{i,j}, r, \alpha, \rho)$ for one 
            \State epoch 
            \Comment{$c$ is the number of classes}
            
            \EndFor
            \EndFor

        \end{algorithmic}
        \label{algo: complete algorithm}
\end{algorithm}


\section{Experiments}

We evaluated our method on three common SSL benchmarks, see Table~\ref{table:datasets}. Unless stated otherwise, the experiments were performed with FixMatch as the SSL module. 
\begin{table}[ht]
\footnotesize
\vspace{-0.5cm}
\begin{center}
\begin{tabular}{l|ccc}
\thead{\bf Name} & \thead{\bf Classes} & \thead{\bf Train/Test Size} & \thead{\bf Dimension}\\
\hline
\rule{0pt}{3ex}\hspace{-.5ex} CIFAR-10 \cite{cifar10}       & 10    & 50000 / 10000   & \small$32\times32\times3$\\
\rule{0pt}{3ex}\hspace{-.5ex} SVHN \cite{svhn}       &  10   &   73257 / 26032 & \small$32\times32\times3$\\
\rule{0pt}{3ex}\hspace{-.5ex} STL-10 \cite{stl10}       & 10   & 5000 / 8000   & \small$96\times96\times3$
\end{tabular}
\end{center}
\vspace{-0.5cm}
\caption{Datasets used in our experiments. For STL-10, the 100K extra unlabeled images were used as well.}
\label{table:datasets}
\end{table}

\subsection{Implementation Details}
\label{subsec: implementation details}

In all of the experiments we used the WideResNet (WRN) architecture \cite{wrn}, replicating the setup described in \cite{fixmatch}. More specifically, for the CIFAR-10 and SVHN datasets we used WRN-28-2, and for STL-10 we used WRN-37-2. In the SSL phase, we kept the exact same hyper-parameters as in the original SSL algorithm being employed, while for the clustering phase, the learning rate and weight decay were reduced to 0.01 and 0.0001 respectively (from 0.03 and 0.0005 in FixMatch). The clustering hyper-parameter $\rho$ was set to 0.2, and the $\alpha$ hyper-parameter was set to 1 for Cifar-10, and 0.6 for SVHN and STL-10. As in most other contemporary SSL methods, we stored and evaluated the model with exponential moving average of the weights over the training and a decay of 0.999.

\textbf{Image augmentation:} during the SSL phase, we took care to always apply the exact same augmentations as used in the original SSL method used to realize the SSL phase. Specifically in the experiments with FixMatch, we used Control Theory Augment \cite{remixmatch} that achieved the best results in most scenarios. In the unsupervised phase, we used the customary flip followed by crop, after the application of random color jitter to each pixel. We used the same flip and crop transformation in both phases: horizontal flip with probability 0.5, followed by cropping the mirror padded image to the original size.  

Unless otherwise mentioned, we trained our model for 200 iterations, each comprising multiple passes over the data in the SSL phase (10 for Cifar-10 and 5 for all other datasets), followed by one pass in the clustering phase.


In all the experiments whose results are reported below, in order to ensure a fair comparison, we used the exact same partitions into labeled and unlabeled data as used in \cite{fixmatch}. Therefore, whenever we present results while replicating experiments reported in \cite{fixmatch}, we use the results reported there for all the methods but our own. When using different existing algorithms, we first made sure that our implementation of those methods yielded comparable results to those reported in the original manuscripts, in order to avoid running all the various experiments anew.

\subsection{Results} \label{subsec: results}

\begin{figure*}[ht]
\includegraphics[width=8cm]{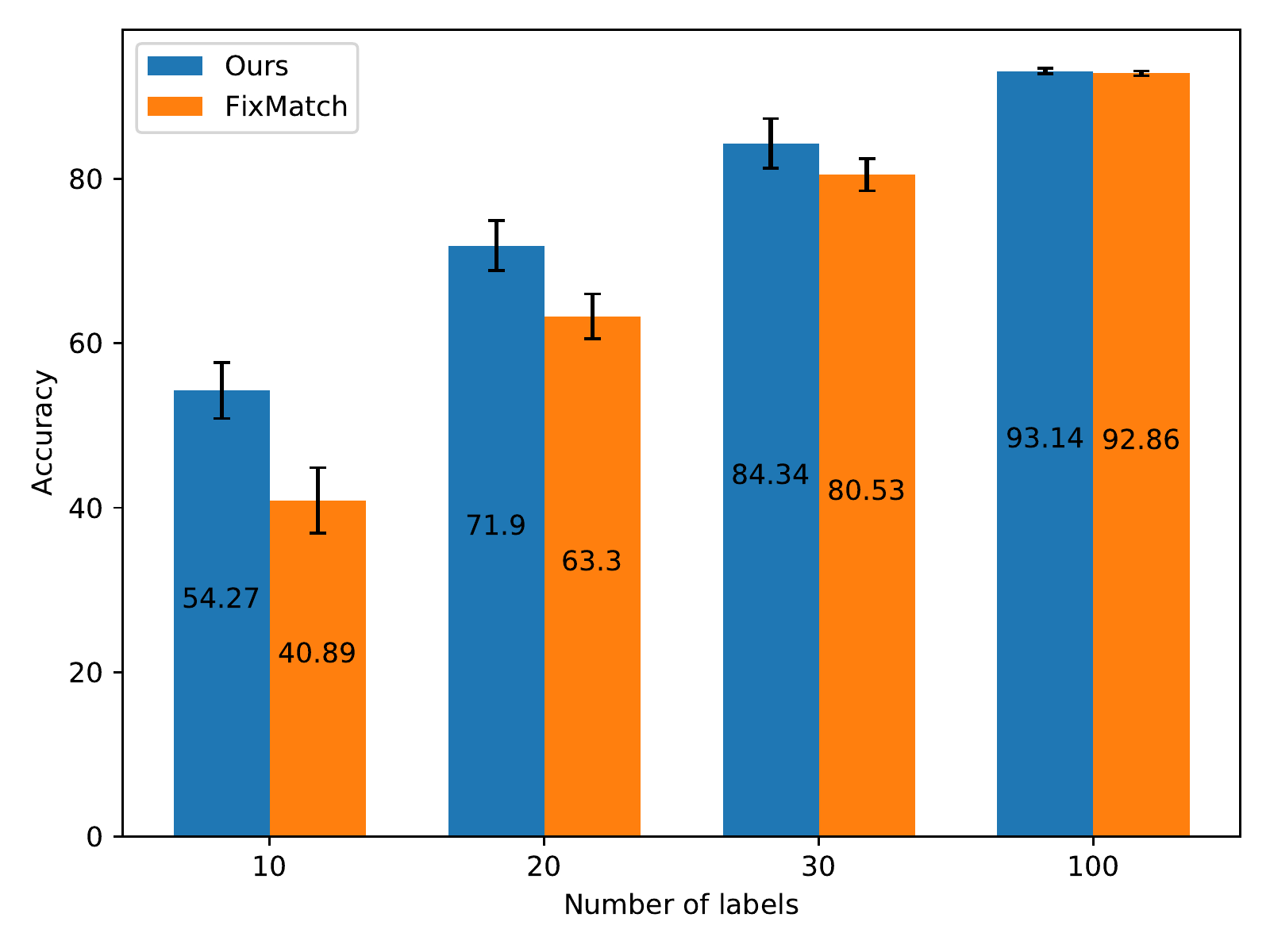}
 \hfill
\includegraphics[width=8cm]{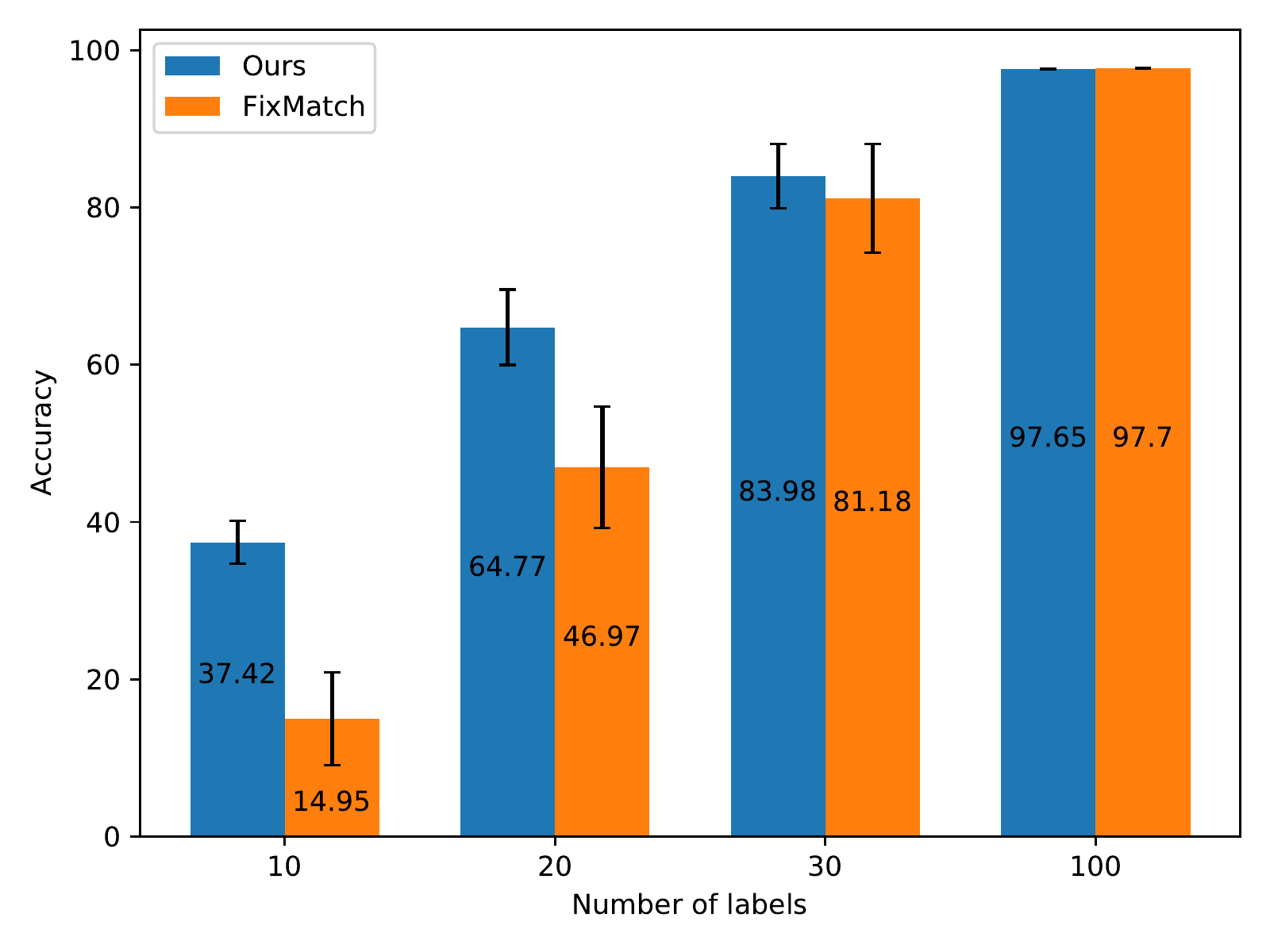}
    \caption{Classification accuracy, comparing our method to FixMatch with CTA, using identical protocols. Left: CIFAR-10, right: SVHN. The numbers inside the columns denote the mean accuracy across different partitions and runs.}
    \label{fig: cifar10 svhn bar graph}
\end{figure*}

\begin{table*}[t]

\begin{center}
\begin{tabular}{l|ccccc}
\toprule
 \multicolumn{1}{c}{} & \multicolumn{2}{c}{CIFAR-10} & STL-10 & \multicolumn{2}{c}{SVHN}\\
\midrule
Method & 40 labels & 250 labels & 1000 labels & 40 labels & 250 labels \\
\midrule
$\Pi$-\text{Model} & - & $54.26 \pm 3.97$ & $26.23 \pm 0.82$ & - & $18.96 \pm 1.92$ \\ 
Pseudo-Labeling & - & $49.78 \pm 0.43$ & $27.99 \pm 0.80$ & - & $20.21 \pm 1.09$ \\
Mean Teacher & - & $32.32 \pm 2.30$ & $21.43 \pm 2.39$ & - & $3.57 \pm 0.11$ \\
MixMatch & $47.54 \pm 11.50$ & $11.05 \pm 0.86$ & $10.41 \pm 0.61$ & $42.55 \pm 14.53$ & $3.98 \pm 0.23$\\
UDA & $29.05 \pm 5.93$ & $8.82 \pm 1.08$ & $7.66 \pm 0.56$ & $52.63 \pm 20.51$ & $5.69 \pm 2.76$ \\
ReMixMatch & $19.10 \pm 9.64$ & $5.44 \pm 0.05$ & $5.23 \pm 0.45$ & $3.34 \pm 0.20$ & $2.92 \pm 0.48$ \\
FixMatch (RA) & $13.81 \pm 3.37$ & $\mathbf{5.07} \pm 0.65$ & $7.98 \pm 1.50$ & $3.96 \pm 2.17$ & $2.48 \pm 0.38$ \\
FixMatch (CTA) & $11.39 \pm 3.35$ & $\mathbf{5.07} \pm 0.33$ & $5.17 \pm 0.63$ & $7.65 \pm 7.65$ & $2.64 \pm 0.64$ \\
\midrule
Ours & $\mathbf{7.39} \pm 0.61$ & $5.51 \pm 0.25$ & $\mathbf{4.78} \pm 0.29$ & $\mathbf{3.09} \pm 0.54$ & $\mathbf{2.30} \pm 0.03$ \\
\bottomrule

 \addlinespace

\end{tabular}
  \caption{Error rates for the 3 datasets used in our study: CIFAR-10, STL-10 and SVHN. Results are reported for varying amounts of labels, denoting the total number of labeled points from all classes.} 
\label{table: cifar-10 and svhn results}
\end{center}
\end{table*}

\subsubsection*{Classification Results with FixMatch}

In Table~\ref{table: cifar-10 and svhn results}, we report the results of our method when applied to the three datasets used in our study, with various amounts of labels. We ran the algorithm with 5 different partitions, the exact same partitions used in \cite{fixmatch}. Due to the large variance in the 40 labels setting, we repeated the experiment 3 times for each partition. Hence, the standard deviation reported in our results is the standard deviation (STD) of the means over the different partitions. As expected for such small partitions, it is rather large.

As can be seen in Table~\ref{table: cifar-10 and svhn results}, our method is very effective in the very small sample regime with 4 labels per class, where its relative advantage over the alternative methods is quite high. Its added value is less pronounced when using a total of 250 labels. Still, when learning to classify the more challenging STL-10 dataset with a setting identical to the one described in~\cite{fixmatch}---we used the same 5 folds of 1000 labeled images each, and the additional 100K unlabeled images---our method once again outperforms all the results reported in \cite{fixmatch}. 

\begin{figure}[ht]
\begin{center}
\includegraphics[width=0.9\columnwidth]{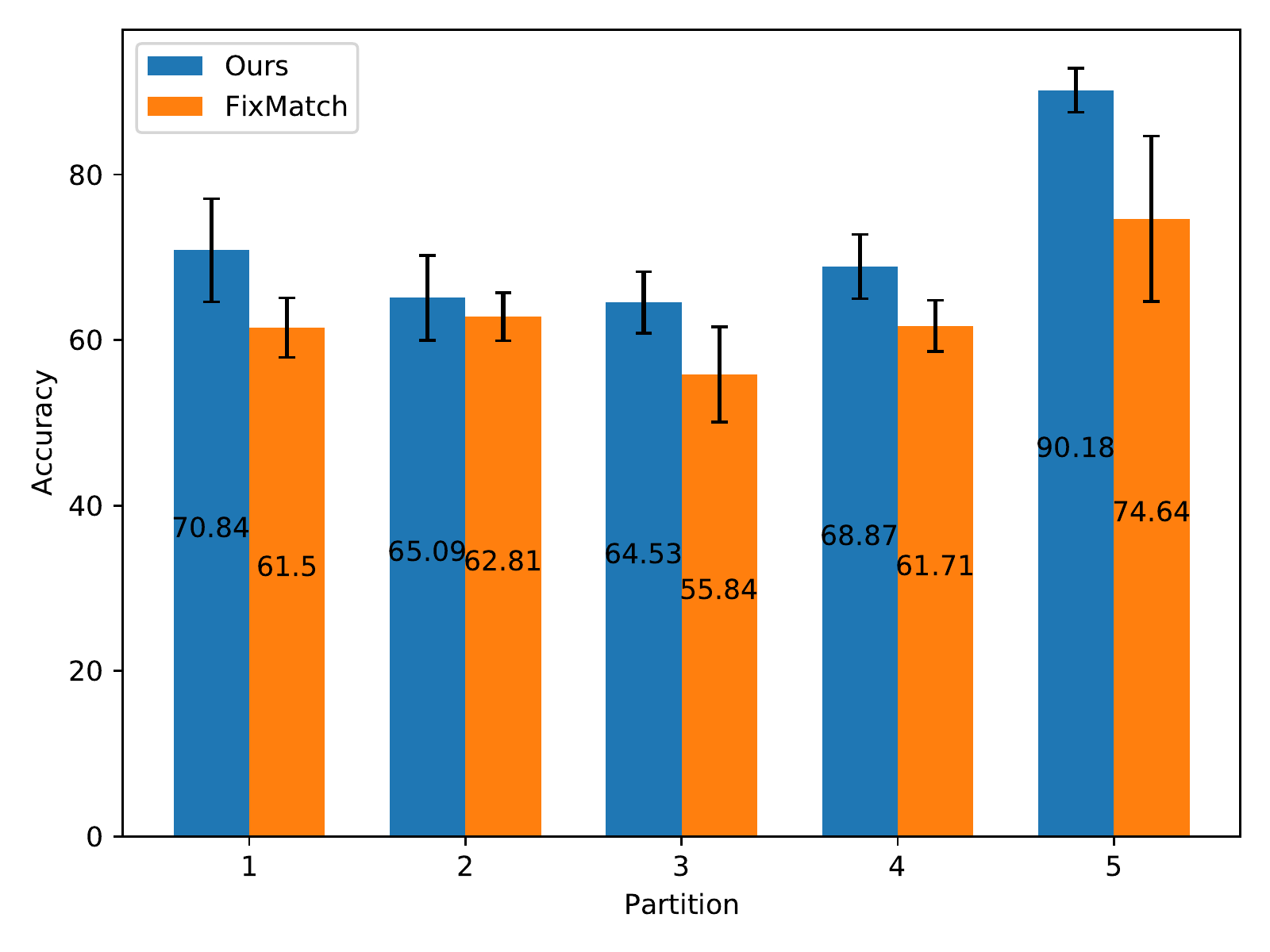}
 \hfill
    \caption{The classification accuracy for CIFAR-10 with 20 labels (2 per class), presented separately for each partition.}
    \label{fig: cifar-10 20 labels partitions bar graph}
\end{center}
\end{figure}

Finally, in order to push our method to its limit, we experimented with even fewer labels: 10, 20, 30, and also 100 - an intermediate number between 40 and 250. This very small sample regime is not systematically investigated in \cite{fixmatch}. The results for these experiments are presented in Fig.~\ref{fig: cifar10 svhn bar graph}, as well as a detailed per-partition results for the 20-labels CIFAR-10 experiment in Fig.~\ref{fig: cifar-10 20 labels partitions bar graph}. Clearly, as long as the number of labeled points is smaller than 250, our method is quite beneficial.

\subsubsection*{Clustering Accuracy Score}

\begin{figure}[ht]
\begin{center}
\includegraphics[width=0.9\columnwidth]{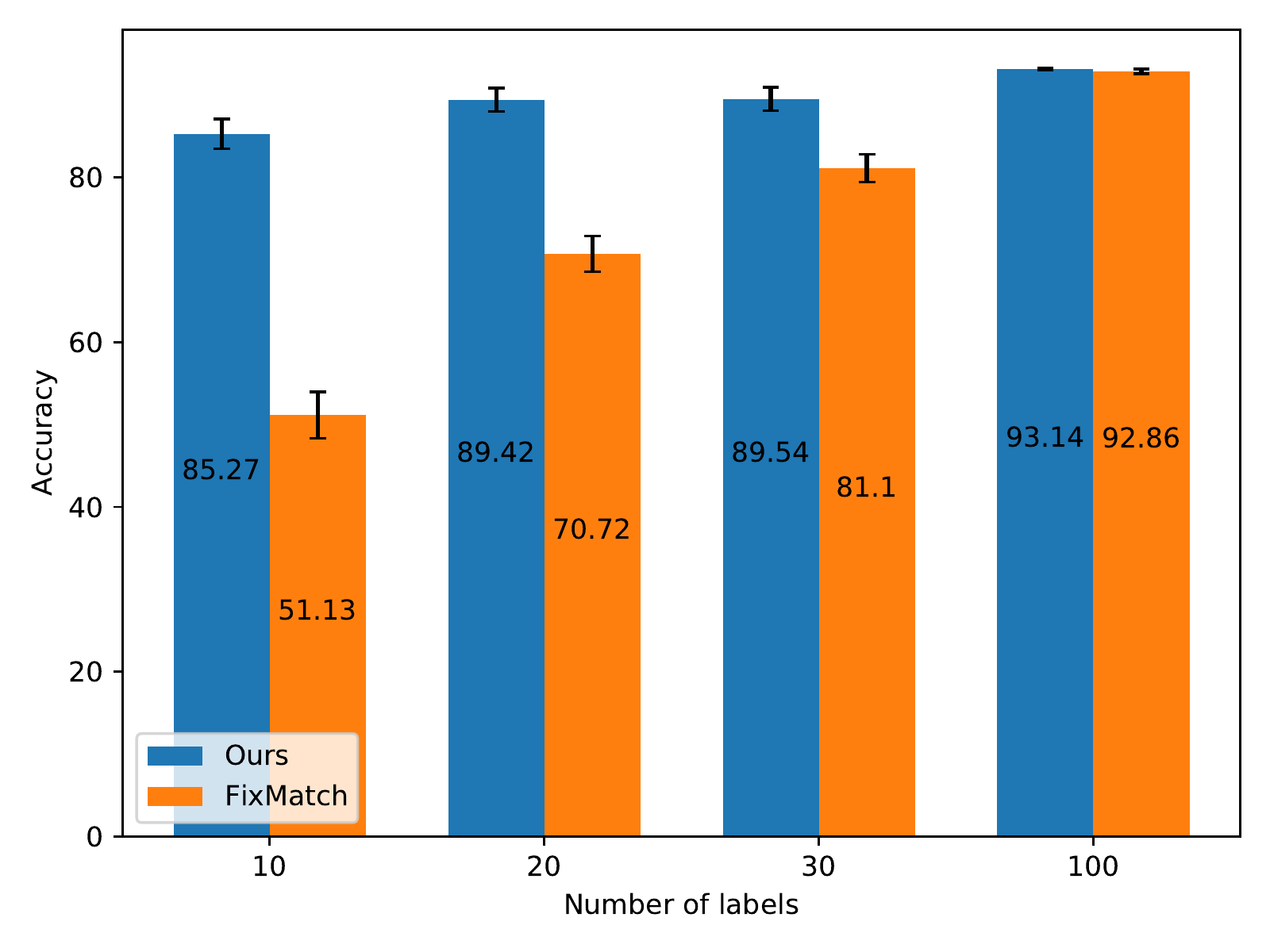}
    \vspace{-0.35cm}
    \caption{Clustering accuracy for the same experiment, the results of which are reported in Fig.~\ref{fig: cifar10 svhn bar graph}.}
    \label{fig: cifar-10_clustering - bar graph}
    \vspace{-0.5cm}
\end{center}
\end{figure}

An interesting advantage of our method is demonstrated in Fig.~\ref{fig: cifar-10_clustering - bar graph}. The accuracy shown there is the clustering accuracy score, which is traditionally computed as the classification accuracy of the best permutation of class labels, when uniquely matched with the different clusters. More precisely, if our test images and their corresponding labels are given by $(\{x_i\}^m_{i=1}, \{y_i\}^m_{i=1})$, and the model's predictions are given by $\{\tilde{y}_i\}^m_{i=1}$, then the clustering accuracy score is defined as:

\begin{equation}\label{equation: clustering accuracy score}
\max_{P \colon [c] \to [c]}\frac{1}{m}\sum_{i=1}^{m}\mathbbm{1}(y_i=P(\tilde{y_i})),
\end{equation}
where $P$ denotes a permutation over the $c$ possible classes.

Note that while the classification accuracy for the experiments with 10 and 20 labels may seem low, the data is still clustered very well by our model. Even with one label per class, the model reaches a mean clustering accuracy of over 85\%, and the best partition achieves mean accuracy of over 90\%. At the same time, we see in Fig.~\ref{fig: cifar10 svhn bar graph} that the mean classification accuracy is only 54.3\%. The gap in accuracy may be large, but it resides solely in the naming of the clusters. Conversely, this cannot be said about the predictions obtained by FixMatch alone. There, the gap between classification accuracy and clustering accuracy is considerably smaller, which means that FixMatch doesn't succeed in separating the classes in the extreme small sample regime. With 100 labeled examples, the classification and clustering accuracy converge to the same value for both methods.

\begin{figure}[ht]
\begin{center}
\includegraphics[width=1.0\columnwidth]{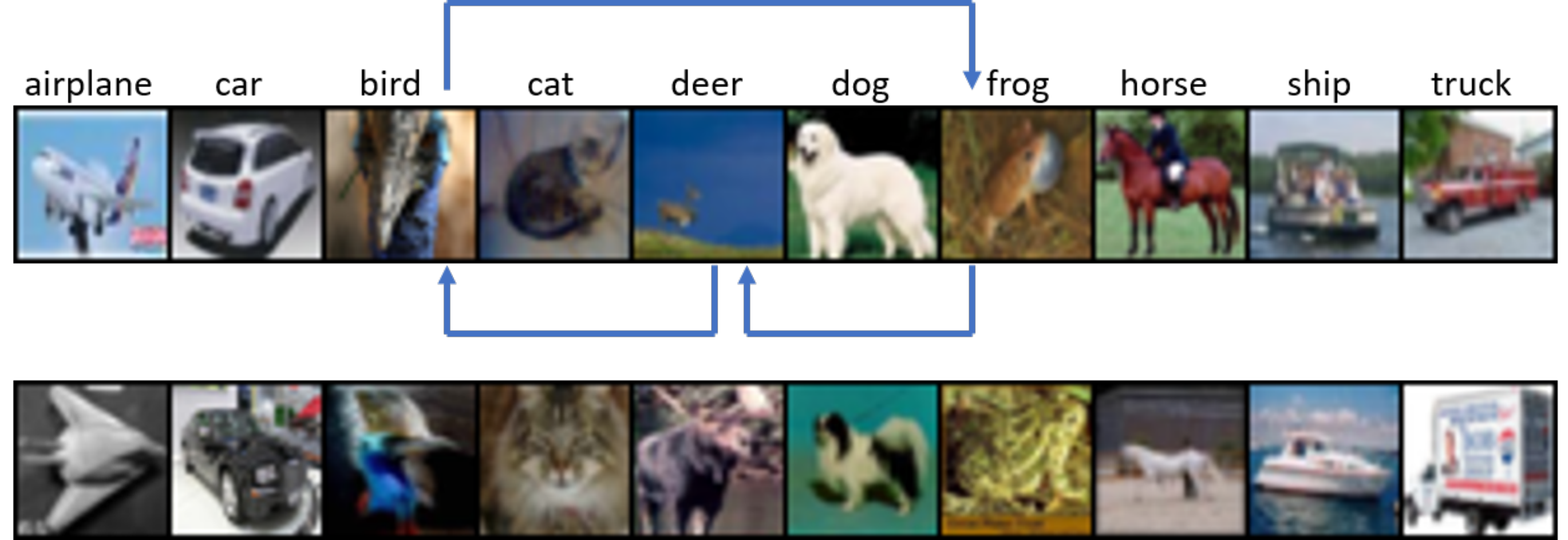}
 \hfill
    \caption{Two different partitions of CIFAR-10 with one label per class are shown above. Given the top partition, our model found the wrong permutation as illustrated with blue arrows. This error led to the gap between classification accuracy of 67.6\% and clustering accuracy of 92.5\%. Given the bottom partition, our model found the optimal permutation and achieved 91.0\% accuracy in both classification and clustering.}
    \label{fig: permutations image}
\end{center}
\end{figure}

In Fig.~\ref{fig: permutations image} we show two partitions from CIFAR-10, each with 10 labels. In one partition there is a big gap between classification accuracy and clustering accuracy, because the model confused the labels of 3 clusters as shown by the arrows. In the other partition, the model succeeded in finding the right permutation, and hence achieved 91\% accuracy in both classification and clustering.

\begin{figure}[ht]
\begin{center}
\includegraphics[width=0.9\columnwidth]{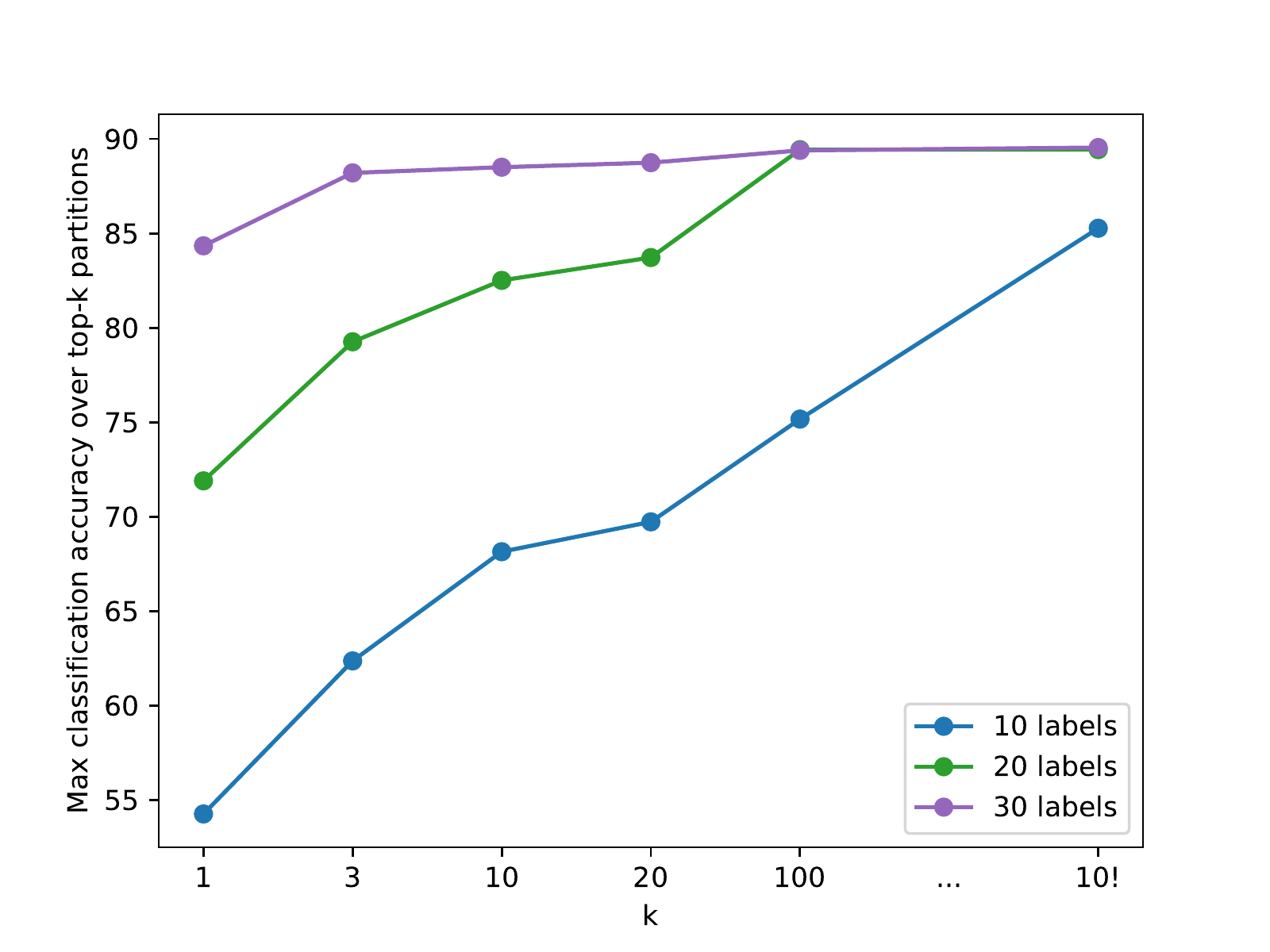}
 \hfill
\vspace{-0.35cm}
    \caption{Top-k classification accuracy for CIFAR-10 with 10, 20 and 30 labels. k denotes the number of permutations considered.}
    \label{fig: top-k permutation accuracy}
\end{center}
\end{figure}

Bridging this gap between classification accuracy and clustering accuracy with so few labels is a hard problem. We investigated a few heuristics in order to identify "good" permutations during or after training, but more effort is needed. In Fig.~\ref{fig: top-k permutation accuracy} we show the results when employing one such heuristic. After training is completed, we rotate the labeled images (in four orientations as in RotNet) and use the average prediction in order to find the $k$ best permutations with Murty's algorithm \cite{murty}. We use rotated images because the trained model is already (over-)fitted to the small labeled sample. Note that the permutation which achieves the best performance, and which defines the clustering accuracy, always lies within the 100 best permutations (out of $10!$ possible permutations) according to this score in the experiments with 20 and 30 labeled examples.


\subsubsection*{Running Time Comparison}

Another advantage of our method is its efficiency with respect to running time. As mentioned in Section~\ref{subsec: implementation details}, the reported results are obtained after 200 iterations of our method. This implies 2000 passes through the whole data, plus another 200 passes for clustering and 200 passes for RotNet. In FixMatch, with randomly sampled batches, the total number of semi-supervised batches processed to achieve the published results approaches ${\sim}1M$ batches, while our method processes ${\sim}220K$ batches to achieve the results shown above. Even though each clustering epoch takes a bit longer than a FixMatch epoch, due to the expensive assignment problem, the total running time of our method adds up to roughly 30\% of the running time needed by FixMatch when left on its own. Similar observations hold for the other SSL algorithms, which are compared with our method next.

\subsubsection*{Other SSL algorithms} 

As explained in Section~\ref{sec:method}, our approach is general in the sense that it can use any clustering algorithm and any SSL method to address the challenging SSL problem of classification with small labeled sample. In this section we show that interlacing our proposed clustering method with two other successful SSL methods improves their outcome, in a similar way to the previous results with FixMatch. Thus, in Table~\ref{table: uda} we show how our approach boosts the performance of MixMatch \cite{mixmatch} and UDA \cite{uda}. These experiments were conducted on CIFAR-10 with 40 labels, using the same 5 partitions as in all the other experiments. Each partition was evaluated once. As can be seen, UDA combined with our clustering mechanism almost closes its initial gap against FixMatch, when both methods use RandAugment as the generator of strong augmentations (this was the augmentation used by the original method). 

\begin{table}[ht]
\begin{center}
\begin{tabular}{|l|c|}
\hline
Method & Error Rate \\
\hline\hline
MixMatch & $47.54 \pm 11.50$\\
MixMatch + Clustering & $\mathbf{35.37} \pm 6.01$\\
\hline\hline

UDA & $29.05 \pm 5.93$\\
UDA + Clustering & $\mathbf{14.37} \pm 3.85$\\
\hline
\end{tabular}
\end{center}
\vspace{-0.1in}
\caption{Error rates of MixMatch (top) and UDA (bottom), with and without clustering, trained on CIFAR-10 with 40 labeled examples.}
\label{table: uda}
\end{table}

\subsection{Ablation Study}

We test the contribution of two major components of the proposed method: clustering and RotNet, using CIFAR-10 with 40 labeled examples. Table \ref{table: ablation} summarizes the results. As before, each of the 5 partitions used in the initial experiments is learned 3 times independently. We can see that RotNet alone does not improve the results, nor does it degrade them. However, without RotNet the clustering phase is far less stable, with a detrimental effect on the final classification outcome.
In both cases, we observed a much higher variance in performance.

\begin{table}[ht]
\begin{center}
\begin{tabular}{|l|c|}
\hline
Method & Error Rate \\
\hline\hline
FixMatch & $11.39 \pm 3.35$ \\
FixMatch + RotNet & $11.55 \pm 2.98$ \\
FixMatch + Clustering & $12.15 \pm 3.08$ \\
Our Method & $\mathbf{7.39} \pm 0.61$ \\
\hline
\end{tabular}
\end{center}
\vspace{-0.2in}
\caption{Ablation study on CIFAR-10 with 40 labeled examples.}
\label{table: ablation}
\end{table}


\section{Summary and Discussion}

Motivated by the desire to reduce the reliance on annotated data as much as possible, we propose a new approach to semi-supervised learning, which is designed to reduce overfit when very few labels are available. The proposed method alternates between an unsupervised clustering phase that ignores the labels in the training data, and a semi-supervised classification phase that makes full use of the training labels. To this end, we propose a new deep clustering algorithm. We then demonstrate the effectiveness of the general approach by plugging into it existing SSL algorithms, achieving significantly improved performance and reduced running time. When the recent FixMatch algorithm is plugged in as the SSL module, we improve state-of-the-art results on 3 benchmarks typically used to evaluate SSL algorithms. The proposed approach is general, in that both the SSL and the clustering modules can be replaced, although in this work we experimented with a single clustering method.

From a broader perspective, our approach can take advantage of curriculum learning \cite{curriculum_learning}, as one might look for means to schedule the supervised and unsupervised phases in a more sophisticated manner during training. It can also benefit from active learning \cite{active_learning}, when seeking the best permutation between labels and clusters in the course of learning, which is especially tricky when the number of labeled points per class is very small. Under the active learning framework, where the learner can opt for a specific label of interest, we can adjust the permutation gradually by clustering the data first, subsequently asking the user to provide labels for each cluster's centroid. This way, in each round of communication the permutation can be tuned with less than one additional label per class, as only labels from uncertain clusters will be requested. 


{\small
\bibliographystyle{ieee_fullname}
\bibliography{egbib}
}

\end{document}